\documentclass[10pt,journal,compsoc]{IEEEtran}
\usepackage{times}  %
\usepackage{mathptmx}  %

\usepackage{tabularx}
\usepackage{multirow} 
\usepackage{threeparttable}
\usepackage{float}
\usepackage{graphicx}
\usepackage{subfigure}
\usepackage{amssymb}
\usepackage{booktabs} 
\usepackage{siunitx} 
\usepackage{graphicx} 
\usepackage{times}
\usepackage{soul}
\usepackage{url}
\usepackage[hidelinks]{hyperref}
\usepackage[utf8]{inputenc}
\usepackage[small]{caption}
\usepackage{enumitem}
\usepackage{graphicx}
\usepackage{amsmath}
\usepackage{amsthm}
\usepackage{booktabs}
\usepackage{algorithm}
\usepackage{algorithmic}
\ifCLASSOPTIONcompsoc
  \usepackage[nocompress]{cite}
\else
  \usepackage{cite}
\fi
\ifCLASSINFOpdf
\else
\fi
\begin{document}
%
\title{QoSBERT: A Uncertainty‑Aware Approach based on Pre‑trained Language Models for Service Quality Prediction}
%
%
%
%

\author{Ziliang~Wang, Xiaohong~Zhang, Ze Shi Li and Meng~Yan, Member, IEEE
\IEEEcompsocitemizethanks{


\IEEEcompsocthanksitem  

Ziliang Wang is with Key Laboratory of High Confidence Software Technologies (Peking University), Ministry of Education; School of Computer Science, Peking University, Beijing, China\\
Xiaohong~Zhang, Meng~Yan are with Key Laboratory of Dependable Service Computing in Cyber Physical Society (Chongqing University),  Ministry of Education, China and School of Big Data and Software Engineering, Chongqing University, Chongqing 401331, China. \protect
\\
Ze Shi Li is in the Department of Computer Science at the University of Victoria, Canada
\IEEEcompsocthanksitem 
Ziliang Wang is the corresponding authors.\\
E-mail: wangziliang@pku.edu.cn
}
}
\IEEEtitleabstractindextext{%
\begin{abstract}
Accurate prediction of Quality of Service (QoS) metrics is fundamental for selecting and managing cloud-based services. 
Traditional QoS models rely on manual feature engineering and yield only point estimates, offering no insight into the confidence of their predictions. 
In this paper, we present QoSBERT, a novel end‑to‑end framework that leverages pre‑trained language models to automatically encode rich, natural‑language descriptions of services and integrates a Monte Carlo Dropout–based uncertainty estimation mechanism to quantify predictive confidence. 
QoSBERT applies attentive pooling over contextualized embeddings and a lightweight multi‑layer perceptron regressor, fine‑tuned jointly to minimize absolute error. We further exploit the resulting uncertainty estimates to select high‑quality training samples, improving robustness in low‑resource settings.
On standard QoS benchmark datasets, QoSBERT achieves an average reduction of 11.7\% in MAE and 6.7\% in RMSE for response time prediction, and 6.9\% in MAE for throughput prediction compared to the strongest baselines, while providing well-calibrated confidence intervals for robust and trustworthy service quality estimation.
Our approach not only advances the accuracy of service quality prediction but also delivers reliable uncertainty quantification, paving the way for more trustworthy, data‑driven service selection and optimization.
\end{abstract}

}

\maketitle

\IEEEdisplaynontitleabstractindextext

%
\IEEEpeerreviewmaketitle

\IEEEraisesectionheading{\section{Introduction}\label{sec:introduction}}
\IEEEPARstart{I}{n} recent times, various domains and applications, such as cloud services, online streaming, and e-commerce, are increasingly being offered as a service, resulting in a greater emphasis on ensuring optimal Quality of Service (QoS)~\cite{muslim2022s,zheng2020web,wang2024deepscaling}.
QoS values are frequently used as crucial inputs for various downstream service computing tasks, such as service recommendation~\cite{liu2019personalized,yao2014unified} and service composition ~\cite{gavvala2019qos,sefati2021qos}.
At the same time, some microservice optimization techniques require accurate predictive Quality of Service (QoS) models, such as autoscaling based on QoS~\cite{47,hussain2022new}.
Consequently, accurately predicting QoS has become a critical topic in service computing ~\cite{shao2007personalized,liu2019context,ghafouri2020survey}.

Existing QoS prediction methods can be broadly categorized into two major paradigms: collaborative filtering-based approaches and deep learning-based approaches. Collaborative filtering methods are typically further divided into memory-based and model-based techniques. 
Memory-based approaches rely on historical QoS interaction data, computing similarity measures among users or services to form neighborhood-based predictions for unobserved QoS values~\cite{zou2018qos,shao2007personalized,tang2012location}.
Despite their straightforward implementation, memory-based approaches often encounter significant limitations due to data sparsity. 
On the other hand, model-based collaborative filtering addresses sparsity by deriving latent semantic representations from QoS interaction histories to identify underlying relationships between users and services~\cite{chen2017exploiting}. 
However, traditional model-based techniques generally capture only linear or simple non-linear interactions, limiting their effectiveness in capturing complex and high-dimensional semantic relationships inherent in QoS data. 
To address this issue, recent studies have explored advanced deep learning techniques~\cite{zou2022ncrl,gao2019context}, utilizing neural network architectures such as multilayer perceptrons to model sophisticated, nonlinear user-service interactions, substantially improving representational richness and predictive accuracy compared to conventional collaborative filtering approaches.

However, accurately predicting QoS is significantly hindered by two major challenges: the underutilization of semantic-rich service and user descriptions, and the lack of uncertainty estimation, both of which limit the robustness and expressiveness of existing models in real-world scenarios.

First, although modern service platforms often provide rich textual descriptions of user contexts and service functionalities, most QoS prediction models rely on sparse numerical features or interaction matrices, failing to capture the inherent semantics embedded in these descriptions~\cite{zou2022ncrl,zhang2019location,lu2023feature,wang2021hsa}.
As a result, existing models often struggle to generalize across heterogeneous services or users, especially when faced with cold-start scenarios or sparse access records. 
For example, zhang et al. encoded the user location information into an integer value to encode the user location information through the encoding layer~\cite{zhang2019location}.

Second, current models typically generate deterministic QoS predictions without quantifying prediction confidence, making it difficult to distinguish between reliable and potentially risky estimates.
This limitation poses a serious concern for practical service selection and adaptation tasks, where accurate confidence estimation is essential for mitigating risks, especially under distribution shifts or noisy inputs. Ye et al. proposed CMF to discuss the study of data noise in QoS tasks, but the research on the trustworthiness of prediction models is still blank~\cite{77}.
Therefore, there is a pressing need to explore models that can (1) effectively encode natural language representations of services and users, and (2) produce well-calibrated QoS predictions with uncertainty awareness.

To address these challenges, we propose QoSBERT, a novel approach that leverages pre-trained language models and uncertainty estimation to improve the accuracy and robustness of QoS prediction.
QoSBERT introduces three innovative components to tackle the aforementioned limitations:
(1) a semantic-aware representation framework based on pre-trained language models,
(2) a Monte Carlo dropout mechanism for predictive uncertainty estimation, and
(3) an attention-enhanced pooling module for refined feature aggregation.

QoSBERT fills the gap in this field by proposing a prediction architecture based on a pre-trained model. First,unlike conventional methods that rely on sparse numerical features or handcrafted vectors, QoSBERT converts user and service attributes into natural language descriptions and encodes them using powerful pre-trained models (e.g., Qwen2.5\cite{yang2024qwen2}).
This enables the model to capture rich semantic features embedded in service texts and user contexts, allowing for better generalization across different domains, even with limited numerical inputs. This has the intuitive benefit that features that were previously difficult to model numerically can be exploited.

Second, to enhance prediction reliability, QoSBERT introduces a Monte Carlo dropout mechanism during inference.
By stochastically sampling multiple forward passes through dropout-enabled layers, the model is able to estimate the variance of predictions and quantify the epistemic uncertainty inherent in sparse QoS datasets.
This uncertainty signal is particularly useful for identifying low-confidence predictions and mitigating overconfident failures. 
Third, we integrate a lightweight attention-based pooling module that fuses token-level representations and enables the model to adaptively focus on semantically critical components in the input sequence.This further boosts the model’s ability to learn fine-grained features relevant to QoS outcomes, even under noisy or long-tail distributed conditions. 

In summary, this paper makes the following key contributions:

a) We propose QoSBERT, the first framework that reformulates QoS prediction as a semantic regression task using pre-trained language models, allowing effective encoding of descriptive user-service information without requiring explicit numerical feature engineering.

b) We design a Monte Carlo dropout-based uncertainty estimation mechanism that enables the model to quantify predictive confidence and improves robustness under sparse or noisy conditions.

c) We conduct extensive experiments on standard QoS benchmarks (e.g., WS-DREAM) and demonstrate that QoSBERT consistently outperforms state-of-the-art baselines, achieving an average reduction of 11.7\% in MAE and 6.7\% in RMSE for response time prediction, and 6.9\% in MAE for throughput prediction. In addition, QoSBERT offers well-calibrated uncertainty estimates, enhancing the interpretability and reliability of QoS-aware decision making.

The remainder of this paper is organized as follows:
Section 2 reviews related work on QoS prediction, with emphasis on collaborative filtering, deep learning, and federated approaches.
Section 3 presents the proposed QoSBERT framework in detail, including its semantic-aware feature construction, transformer-based encoding, and uncertainty-aware regression head.
Section 4 describes the experimental setup, including dataset configurations, evaluation metrics, baseline models, and implementation details.
Section 5 reports the experimental results and performance comparisons across different QoS metrics and data sparsity levels.
Section 6 concludes the paper and outlines future directions, such as model compression and extension to multi-modal QoS scenarios.

\section{Related work}\label{sec:Related work}
\textbf{Collaborative Filtering for QoS Prediction.}
Collaborative filtering (CF) has been widely employed for Quality of Service (QoS) prediction due to its ability to capture behavioral patterns among users and services based on historical interactions~\cite{1,2,3,4}.
Traditional CF-based methods can be broadly categorized into two families: memory-based and model-based approaches~\cite{13}.
Memory-based CF directly estimates unknown QoS values by measuring similarity between users or services, often relying on features such as geographic location, response time profiles, or network attributes.
Representative examples include user-based methods like UPCC~\cite{12}, service-oriented techniques like IPCC~\cite{13}, and hybrid extensions such as UIPCC~\cite{14} that jointly consider user-service pairs. 
These techniques are computationally efficient and simple to implement, but tend to suffer from poor scalability and limited modeling capacity in sparse environments.

In contrast, model-based CF learns latent representations of users and services through training on historical QoS matrices, enabling better generalization and handling of data sparsity.
Matrix factorization (MF) is one of the most prominent model-based paradigms~\cite{8,9}, where latent vectors are optimized to approximate observed QoS entries. 
Further enhancements integrate auxiliary information such as time~\cite{35}, location~\cite{tang2012location}, or trust relationships to refine prediction.
To address overfitting and interpretability, several variants introduce constraints or regularization; for instance, Luo et al. proposed a non-negative latent factor model (NLF) that incorporates structural assumptions into latent space~\cite{29}.
Despite their predictive power, both memory- and model-based CF approaches often rely on centralized data collection, raising concerns about privacy leakage and making them less suitable for privacy-sensitive or distributed service environments.

\textbf{Deep Learning-Based QoS prediction  Approaches.}
\textbf{Deep Learning for QoS Prediction.}
In recent years, deep learning techniques have been increasingly adopted to address the limitations of traditional collaborative filtering in modeling non-linear user-service relationships.
Neural network-based collaborative filtering (NCF)~\cite{20} pioneered this direction by replacing the inner product with a trainable neural architecture, allowing more expressive interaction modeling.
To further incorporate temporal and contextual patterns, various time-aware deep models have been proposed.
For instance, Zhou et al.~\cite{83} encoded temporal dynamics by associating latent variables with discrete time intervals, while Wang et al.~\cite{wang2016online} captured sequential dependencies using dynamic Bayesian structures.
Other efforts integrate multiple modalities; DHSR~\cite{xiong2018deep} fused text semantics and neural networks to infer complex relations between services and mashups, improving recommendation performance.
More recently, graph neural networks (GNNs), residual connections, and deep latent state models have been explored to better leverage structured and implicit information in sparse QoS scenarios~\cite{liu2023qosgnn,zhang2021probability,86}.

Beyond modeling accuracy, recent studies have increasingly emphasized trustworthiness and data privacy in QoS prediction. 
For example, FRLN~\cite{zou2024frln} introduces a residual ladder network within a federated learning framework to extract multi-level latent features while safeguarding user data through personalized local training. 
Similarly, PE-FGL~\cite{zou2025privacy} leverages privacy-enhancing graph construction and secure federated aggregation to enable accurate and confidential QoS estimation across distributed clients.

\begin{figure*}
\centering
\includegraphics[width=0.88\textwidth]{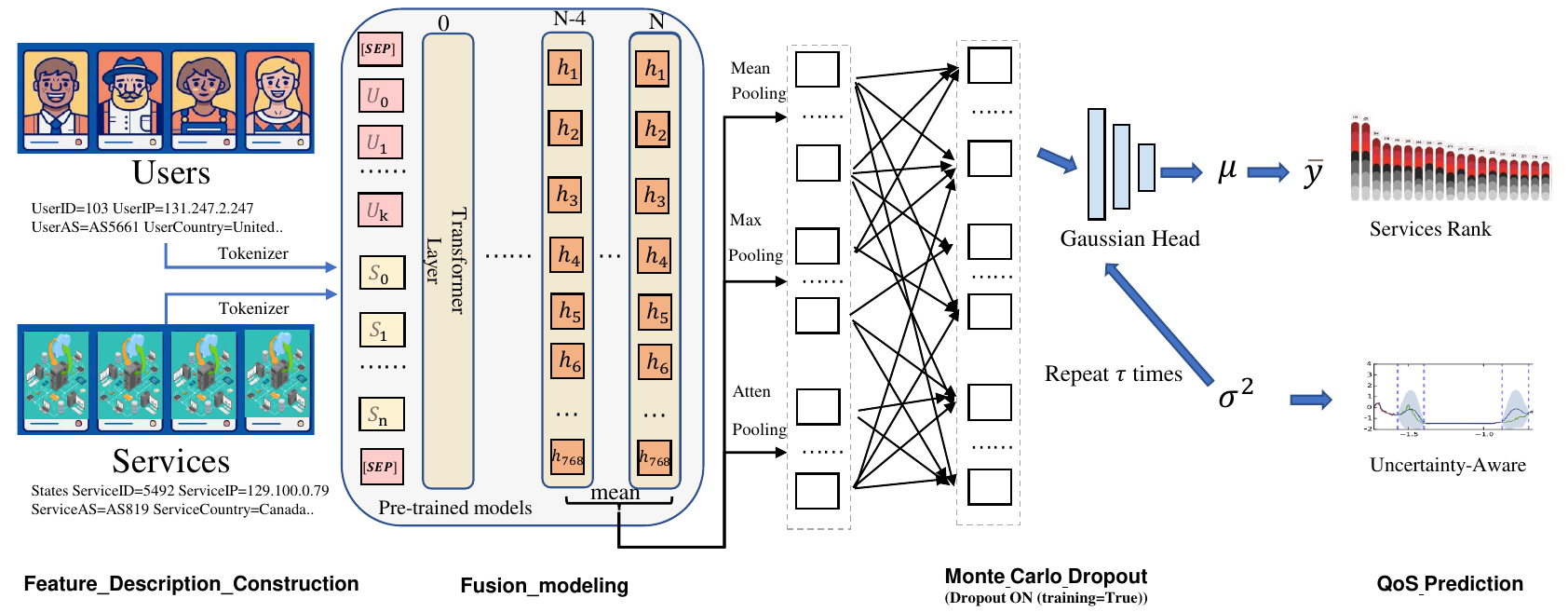}
\captionsetup{justification=centering}
 \vspace{-1em}
\caption{The overall architecture of QoSBERT, which integrates pre-trained language models with multi-head pooling and Monte Carlo dropout to enable uncertainty-aware QoS prediction based on user-service textual descriptions.}
 \vspace{-1em}
\end{figure*}

\section{Approach}
The architecture of QoSBERT, as illustrated in Fig. 2, consists of four primary stages: feature description construction, contextual fusion via pre-trained models, uncertainty-aware prediction using Monte Carlo Dropout, and calibrated confidence estimation via temperature scaling.








\subsection{Feature Description Construction}
\subsubsection{Semantic Feature Encoding from Service and User Metadata}

\indent
Traditional QoS prediction models typically encode users and services using discrete IDs or sparse numerical features, which severely limits their ability to capture the underlying semantics of user-service interactions. These handcrafted or categorical features fail to express contextual relationships such as geographic proximity, network similarity, or routing topologies.

To address this limitation, we propose a semantic feature construction strategy that transforms structured metadata (e.g., IP address, AS number, country, region) into unified natural language descriptions. Specifically, we parse the raw metadata embedded in each QoS record and generate a standardized textual summary using predefined templates.

Each QoS sample is initially defined as a triplet:
\begin{equation}
\centering
L=(U, S, T)
\end{equation}
where $U$ and $S$ represent the user and service metadata respectively, and $T$ is the observed QoS value (e.g., response time or throughput).

We implement a rule-based parser that extracts structured fields—such as \texttt{UserID}, \texttt{IP address}, \texttt{AS number}, and \texttt{Country}—from annotated data, and rephrases them into a unified template-based natural language form. 



This string is then stored under the \texttt{"feature"} field in the JSONL data format, replacing the original raw code or sparse numerical representation. The transformation is implemented through a script that uses regular expressions to extract field values and format them into the template above, as shown in our data preprocessing pipeline.
This textual encoding provides two key advantages:
\begin{enumerate}
    \item It allows seamless integration with pre-trained language models (e.g., RoBERTa~\cite{liu2019roberta}, CodeGen~\cite{nijkamp2022codegen}), which are naturally designed to process tokenized text and leverage semantic priors from large-scale corpora.
    \item It enriches the sparse categorical metadata by projecting it into dense, contextualized embeddings, enabling better generalization in cases such as cold-start scenarios or rare service-user pairs.
\end{enumerate}

These tokenized natural language inputs are then passed to the Transformer encoder for further contextual feature fusion, which we elaborate in the next subsection.

\subsection{Fusion Modeling with Pre-trained Models}
\subsubsection{Contextual Representation}
\begin{algorithm}[t]
\caption{QoSBERT: QoS Prediction with Uncertainty Estimation}
\label{alg:qosbert}
\begin{algorithmic}[1]
\REQUIRE QoS records $L = \{(U_i, S_i, T_i)\}_{i=1}^{N}$; dropout iterations $T$; temperature parameter $\tau$
\ENSURE Predicted QoS $\mu$ and calibrated variance $\sigma^2_{\text{cal}}$

\STATE // Feature Construction
\FOR{$i=1$ to $N$}
    \STATE Generate description $x_i = \texttt{desc}(U_i) \Vert \texttt{desc}(S_i)$
\ENDFOR

\STATE // Contextual Encoding
\FOR{$i=1$ to $N$}
    \STATE Tokenize and encode $x_i$ with pretrained Transformer to get $H_i^{(1)}, ..., H_i^{(L)}$
    \STATE Fuse top-$K$ layers: $H_i^{\text{fused}} = \frac{1}{K} \sum_{l=L-K+1}^{L} H_i^{(l)}$
    \STATE Apply mean, max, and attention pooling:
    \[
    v_i = \text{Mean}(H_i^{\text{fused}}) + \text{Max}(H_i^{\text{fused}}) + \text{Attn}(H_i^{\text{fused}})
    \]
\ENDFOR

\STATE // Monte Carlo Dropout Sampling
\FOR{$t=1$ to $T$}
    \STATE Apply dropout and get $(\mu_i^{(t)}, \sigma_i^{2(t)}) = f_{\theta}^{(t)}(v_i)$
\ENDFOR

\STATE Compute aggregated prediction:
\[
\mu_i = \frac{1}{T} \sum_{t=1}^{T} \mu_i^{(t)}, \quad 
\sigma_i^2 = \frac{1}{T} \sum_{t=1}^{T} \sigma_i^{2(t)}
\]

\STATE // Temperature Scaling (Inference Only)
\[
\sigma_{i,\text{cal}}^2 = \exp(2 \cdot \log \tau) \cdot \sigma_i^2
\]

\RETURN $\mu_i$, $\sigma_{i,\text{cal}}^2$
\end{algorithmic}
\end{algorithm}

Following the construction of semantic feature descriptions, we encode the user-service textual pairs using a pre-trained Transformer model (e.g., CodeGen or Qwen2.5-0.5B-Instruct) to obtain contextualized token representations.
Each textual input is first tokenized into subword units using the corresponding tokenizer of the backbone model, transforming the raw description into a sequence of discrete token IDs.
Formally, given a user description \( U \) and a service description \( S \), the input sequence is structured as:
\[
X = [\texttt{[BOS]}, U_0, ..., U_k, \texttt{[SEP]}, S_0, ..., S_n, \texttt{[EOS]}]
\]
where \texttt{[BOS]} and \texttt{[EOS]} denote the beginning and end of the sequence, and \texttt{[SEP]} separates user and service parts.

The tokenized input \( X \) is then passed through the Transformer encoder, yielding a set of hidden states:
\[
H = \{h_1, h_2, ..., h_L\} \in \mathbb{R}^{L \times d}
\]
where \( L \) is the sequence length and \( d \) is the hidden dimensionality.

\vspace{0.5em}
\textbf{Multi-Layer Feature Fusion.}
Recent empirical studies~\cite{de2020s,liu2019roberta} have shown that different layers in Transformer models encode different levels of linguistic and semantic information.
Lower layers typically capture local syntactic patterns, while deeper layers represent more abstract and task-specific features.
However, relying solely on the last layer may lead to overfitting on noisy or unstable features, especially when the downstream dataset is small.
Therefore, to balance generality and task-awareness, we aggregate the representations from the last \( K \) encoder layers (default \( K=4 \)) to form a fused contextual feature:
\[
H^{\text{fused}} = \frac{1}{K} \sum_{i=N-K+1}^{N} H^{(i)}
\]
where \( N \) is the total number of Transformer layers, and \( H^{(i)} \) is the hidden state output at layer \( i \).

This layer fusion strategy not only smooths potential noise from individual layers but also captures richer semantic cues for QoS prediction.

\vspace{0.5em}

\textbf{(2) Multi-pooling.}
We further apply three pooling strategies on \( H^{\text{fused}} \) to extract global sequence features:
\begin{itemize}
    \item \textbf{Mean Pooling:} element-wise mean over all tokens.
    \item \textbf{Max Pooling:} element-wise maximum over all tokens.
    \item \textbf{Attention Pooling:} a learnable self-attention pooling module to capture task-specific token contributions.
\end{itemize}

The three pooled vectors are added to form the final sequence embedding:
\[
\mathbf{v}_{\text{seq}} = \text{Mean}(H^{\text{fused}}) + \text{Max}(H^{\text{fused}}) + \text{Attn}(H^{\text{fused}})
\]

This vector \( \mathbf{v}_{\text{seq}} \) represents the joint semantics of user and service, and is used as input for the downstream QoS regression module described in the next subsection.

\subsection{Uncertainty-Aware QoS Prediction}

\subsubsection{Monte Carlo Dropout and Variance Estimation}

To estimate both the expected QoS value and its predictive uncertainty, we adopt a probabilistic regression strategy based on Monte Carlo Dropout (MC Dropout)~\cite{gal2016dropout}. During inference, we perform $T$ stochastic forward passes through the model with dropout layers activated, and collect multiple samples of the predicted Gaussian parameters $(\mu^{(i)}, \sigma^{2(i)})$.

Formally, for each pass $i = 1, ..., T$, the regression head outputs:

\[
(\mu^{(i)}, \sigma^{2(i)}) = \mathcal{F}_\theta^{(i)}(\mathbf{v}_{\text{seq}})
\]

The final prediction is obtained by averaging the sampled means and variances:

\[
\mu = \frac{1}{T} \sum_{i=1}^{T} \mu^{(i)}, \quad 
\sigma^2 = \frac{1}{T} \sum_{i=1}^{T} \sigma^{2(i)}
\]

This sampling procedure allows us to capture the epistemic uncertainty associated with the model's predictions.

\subsubsection{Temperature Scaling for Calibration}

To improve the calibration of the predicted uncertainty, we introduce a learnable temperature parameter $\tau$, which scales the estimated variance:

\[
\sigma_{\text{cal}}^2 = \tau^2 \cdot \sigma^2, \quad \tau = \exp(\log \tau)
\]

The temperature parameter $\log \tau$ is optimized on a held-out calibration set by minimizing the Gaussian negative log-likelihood (NLL):

\[
\mathcal{L}_{\text{NLL}} = \frac{1}{N} \sum_{i=1}^{N} \left[ \frac{(y_i - \mu_i)^2}{2\sigma_{\text{cal},i}^2} + \frac{1}{2} \log \sigma_{\text{cal},i}^2 \right]
\]

\subsubsection{Final Objective}

During training, the model jointly minimizes the negative log-likelihood and the Mean Absolute Error (MAE) to balance probabilistic calibration and point-wise accuracy:

\[
\mathcal{L} = \mathcal{L}_{\text{NLL}} + \lambda \cdot \mathcal{L}_{\text{MAE}}, \quad 
\mathcal{L}_{\text{MAE}} = \frac{1}{N} \sum_{i=1}^{N} |y_i - \mu_i|
\]

Here, $\lambda$ is a hyperparameter that controls the trade-off between calibration and prediction precision. At inference time, we report both the predicted mean $\mu$ and calibrated uncertainty $\sigma^2_{\text{cal}}$ to support risk-aware decision making in QoS-sensitive applications.

\subsection{Implementation details}
\subsubsection{Gradual Layer Unfreezing for Representation Refinement}
\label{sec:unfreeze}
\begin{table}[ht]
\centering
\caption{Hyperparameters used for training QoSBERT.}
\begin{tabular}{ll}
\toprule
\textbf{Hyperparameter} & \textbf{Value} \\
\midrule
Model type & Qwen (2.5-0.5B-Instruct) \\
MC Dropout passes ($T$) & 20 \\
Dropout probability & 0.1 \\
Block size & 128 \\
Batch size & 128 \\
Learning rate $\lambda$ & $2 \times 10^{-5}$ \\
Max gradient norm & 1.0 \\
Random seed & 42 \\
\bottomrule
\end{tabular}
\label{tab:qosbert_hyperparam}
\end{table}

To further enhance the learning capability of the encoder and enable more task-specific adaptation, we adopt a \textbf{gradual layer unfreezing} strategy inspired by domain-adaptive pretraining techniques~\cite{wu2021domain}.

Instead of fine-tuning the entire pretrained model at once, which may lead to catastrophic forgetting or overfitting on small-scale QoS data, we initially freeze all encoder parameters except for LayerNorm and Embedding layers. This guarantees stable gradient propagation and preserves general knowledge encoded in the backbone model.

During training, the model progressively unfreezes the top-most $n$ layers of the Transformer encoder in every $k$ epochs, allowing deeper semantic alignment with the QoS prediction task while avoiding early instability. Let $\mathcal{L}_{\text{encoder}} = \{\ell_1, \ell_2, ..., \ell_L\}$ denote the list of all encoder blocks, our approach incrementally activates gradient updates for the top layers:
\begin{equation}
\mathcal{L}_{\text{active}}^{(e)} = \{\ell_{L-n(e)+1}, ..., \ell_{L}\}
\end{equation}
where $n(e)$ is the number of unfreezed layers at epoch $e$, controlled by a linear schedule.

This strategy offers two advantages:
\begin{itemize}
    \item [1)] It enables the model to retain generalizable features from pretraining while gradually adapting to domain-specific patterns in service quality data.
    \item [2)] It stabilizes the optimization landscape in early epochs and improves convergence in the later stage by carefully exposing trainable parameters.
\end{itemize}

\subsection{Complexity Analysis}
\label{sec:complexity}

In this subsection, we analyze the computational complexity of QoSBERT during both training and inference stages.

\vspace{0.5em}
\textbf{Training Complexity.}
During training, each input description is tokenized and fed into a pre-trained Transformer encoder.
The dominant computational cost stems from the Transformer forward pass, which scales as:
\[
\mathcal{O}(L^2 d)
\]
where $L$ is the input sequence length and $d$ is the hidden dimension size.
Following the encoder, the multi-view pooling (mean, max, attention) operations are linear in $L$, and the subsequent regression head introduces negligible additional complexity compared to the encoder backbone.

Moreover, the gradual unfreezing strategy ensures that in the early epochs, only a small subset of layers are updated, effectively reducing training overhead during the initial optimization phase.

\vspace{0.5em}
\textbf{Inference Complexity.}
In the inference phase, QoSBERT applies Monte Carlo Dropout to perform $T$ stochastic forward passes for uncertainty estimation.
Therefore, the total inference cost is approximately $T$ times that of a standard single forward pass:
\[
\mathcal{O}(T \times (L^2 d))
\]
where $T$ is typically set to a moderate value (e.g., $T=20$) to balance between predictive performance and efficiency.

Since only the regression head is repeatedly applied after the encoder outputs are obtained, and considering modern GPU parallelization capabilities, this overhead remains acceptable for most QoS prediction scenarios.

Overall, QoSBERT introduces moderate additional computational overhead compared to standard Transformer-based regression models due to the use of MC Dropout.
However, this overhead is justified by the significant gains in predictive uncertainty estimation and model robustness, which are critical for reliable QoS-sensitive service computing applications.

\section{STUDY SETUP}\label{sec:STUDY SETUP}

\begin{table}[ht]
\centering
\scriptsize
\caption{Division of dataset under different sparsity levels.}
\renewcommand{\arraystretch}{1.2}
\begin{tabularx}{8.5cm}{lllXXX}
\toprule
No. & Density & Split (Train:Test:Valid) & Train & Test & Validation \\
\midrule
RT:D1.1 & 0.05 & 5\%:75\%:20\% & 95,877 & 1,437,361 & 383,310 \\
RT:D1.2 & 0.10 & 10\%:70\%:20\% & 191,755 & 1,341,483 & 383,310 \\
RT:D1.3 & 0.15 & 15\%:65\%:20\% & 287,632 & 1,245,606 & 383,310 \\
RT:D1.4 & 0.20 & 20\%:60\%:20\% & 383,510 & 1,149,728 & 383,310 \\
\midrule
TP:D2.1 & 0.05 & 5\%:75\%:20\% & 82,586 & 1,218,788 & 330,343 \\
TP:D2.2 & 0.10 & 10\%:70\%:20\% & 165,171 & 1,136,203 & 330,343 \\
TP:D2.3 & 0.15 & 15\%:65\%:20\% & 247,757 & 1,053,617 & 330,343 \\
TP:D2.4 & 0.20 & 20\%:60\%:20\% & 330,343 & 971,031 & 330,343 \\
\bottomrule
\end{tabularx}
\label{tab:density_splits}
\end{table}


\subsection{Datasets}
\label{sec:datasets}

We evaluate the proposed QoSBERT model on a refined version of the widely used WS-Dream dataset~\cite{3}.
WS-Dream contains Quality of Service (QoS) records collected from 339 users invoking over 5,800 real-world web services across various geographical regions.
Each invocation record includes two types of QoS attributes: response time (RT, in seconds) and throughput (TP, in kbps).
To better align the data with the modeling assumptions of QoSBERT and enhance prediction reliability, we preprocess the original dataset through the following steps:

\begin{table}[ht]
\centering
\caption{Original Data Structure of QoS Records, Users, and Services.}
\label{tab:original_data}
\renewcommand{\arraystretch}{1.2}
\begin{tabular}{ll}
\toprule
\textbf{Field} & \textbf{Description} \\
\midrule

\multicolumn{2}{l}{\textit{User Metadata}} \\
IP Address & Public IP address of the user. \\
Country & Country derived from IP geolocation. \\
IP Number & Numeric representation of the IP. \\
AS & Registered Autonomous System name and number. \\
Latitude/Longitude & Geolocation coordinates. \\
\midrule
\multicolumn{2}{l}{\textit{Service Metadata}} \\
WSDL Address & URL of the service description (WSDL). \\
Service Provider & Hosting organization. \\
IP Address & Public IP address of the service server. \\
Country & Service server location by country. \\
IP Number & Numeric representation of the service IP. \\
AS & Hosting Autonomous System. \\
Latitude/Longitude & Server geolocation coordinates. \\
\bottomrule
\end{tabular}
\end{table}


\vspace{0.5em}
\textbf{(1) Semantic Feature Construction.}
Each QoS record originally consists of basic invocation information and limited numerical features regarding users and services.
To enable semantic understanding, we enrich these records by retrieving additional metadata from the user and service tables, such as IP addresses, Autonomous Systems (AS), geographic locations, and WSDL providers.
These fields are then assembled into structured natural language templates, which serve as inputs to the Transformer encoder.
This textual construction not only provides rich contextual cues for model learning but also bridges the gap between sparse numerical features and pre-trained language models' requirements.

\vspace{0.5em}
\textbf{(2) Data Sparsity Emulation.}
Table~\ref{tab:density_splits} summarizes the dataset partitioning under varying training densities. 
For both RT (Response Time) and TP (Throughput) tasks, we adopt four levels of data sparsity: 5\%, 10\%, 15\%, and 20\% training density, simulating different levels of data availability in real-world scenarios. 
A fixed 20\% of the total dataset is reserved as the validation set for hyperparameter tuning and model calibration, while the remaining samples are assigned to the test set.
This split strategy ensures a consistent evaluation framework while systematically analyzing the model's robustness under different degrees of data sparsity.



\begin{table*}[ht]
\centering
\caption{Experimental Results of QoS Prediction Under Multiple Densities on RT Dataset}
\label{tab:qosbert_rt_results}
\renewcommand{\arraystretch}{1.4} %
\setlength{\tabcolsep}{12pt}
\begin{tabular}{lcccccccc}
\toprule
\multirow{2}{*}{Methods} & \multicolumn{2}{c}{Density 5\%} & \multicolumn{2}{c}{Density 10\%} & \multicolumn{2}{c}{Density 15\%} & \multicolumn{2}{c}{Density 20\%} \\
\cmidrule(lr){2-3} \cmidrule(lr){4-5} \cmidrule(lr){6-7} \cmidrule(lr){8-9}
& MAE & RMSE & MAE & RMSE & MAE & RMSE & MAE & RMSE \\
\midrule
UPCC  & 0.698 & 1.665 & 0.559 & 1.466 & 0.496 & 1.349 & 0.464 & 1.274 \\
LACF  & 0.631 & 1.439 & 0.562 & 1.338 & 0.513 & 1.269 & 0.477 & 1.222 \\
PMF   & 0.623 & 1.532 & 0.528 & 1.329 & 0.488 & 1.238 & 0.469 & 1.202 \\
NCF   & 0.472 & 1.438 & 0.386 & 1.314 & 0.362 & 1.303 & 0.352 & 1.274 \\

LDCF         & 0.403 & 1.277 & 0.364 & 1.233 & 0.345 & 1.169 & 0.331 & 1.138 \\
FRLN         & 0.379 & 1.306 & 0.344 & 1.238 & 0.322 & 1.201 & 0.309 & 1.175 \\
 PE-FGL         & 0.369 & 1.305 & 0.339 & 1.258 & 0.331 & 1.233 & 0.303 & 1.183 \\
\textbf{QoSBERT} & \textbf{0.327} & \textbf{1.199} & \textbf{0.317} & \textbf{1.210} & \textbf{0.294} & \textbf{1.175} & \textbf{0.249} & \textbf{1.064} \\
\midrule
Gain (\%) & \textbf{+11.38\%} & \textbf{+8.12\%} & \textbf{+6.49\%} & \textbf{+3.82\%} & \textbf{+11.18\%} & \textbf{+4.70\%} & \textbf{+17.82\%} & \textbf{+10.06\%} \\
\bottomrule
\end{tabular}
\end{table*}

\begin{table*}[ht]

\centering
\caption{Experimental Results of QoS Prediction Under Multiple Densities on TP Dataset}
\label{tab:qosbert_tp_results}
\renewcommand{\arraystretch}{1.4} %
\setlength{\tabcolsep}{13pt}
\begin{tabular}{lcccccccc}
\toprule
\multirow{2}{*}{Methods} & \multicolumn{2}{c}{Density 5\%} & \multicolumn{2}{c}{Density 10\%} & \multicolumn{2}{c}{Density 15\%} & \multicolumn{2}{c}{Density 20\%} \\
\cmidrule(lr){2-3} \cmidrule(lr){4-5} \cmidrule(lr){6-7} \cmidrule(lr){8-9}
& MAE & RMSE & MAE & RMSE & MAE & RMSE & MAE & RMSE \\
\midrule
UPCC  & 31.43 & 77.08 & 24.70 & 64.18 & 22.35 & 58.95 & 21.21 & 56.16 \\
LACF  & 22.97 & 55.78 & 19.44 & 52.92 & 17.58 & 49.56 & 16.45 & 47.41 \\
PMF   & 26.47 & 67.46 & 19.83 & 50.64 & 16.84 & 47.48 & 15.32 & 43.86 \\
NCF   & 18.68 & 54.65 & 14.40 & 46.22 & 13.30 & 45.35 & 12.84 & 44.95 \\

FRLN         & 14.94 & 51.62 & 12.57 & 44.67 & 11.37 & 40.87 & 11.06 & 39.60 \\
LDCF         & 13.84 & 47.35 & 12.38 & 43.48 & 11.27 & 39.81 & 10.84 & 38.99 \\
PE-FGL        & 13.36 & 44.71 &11.47& 41.51& 10.35& 37.41 & 9.95 & 36.35 \\
\textbf{QoSBERT} & \textbf{12.61} & \textbf{44.31} & \textbf{10.75} & \textbf{40.54} & \textbf{9.58} & \textbf{35.65} & \textbf{9.14} & \textbf{35.93} \\
\midrule
Gain (\%) & \textbf{+5.61\%} & \textbf{+0.89\%} & \textbf{+6.28\%} & \textbf{+2.23\%} & \textbf{+7.44\%} & \textbf{+4.70\%} & \textbf{+8.14\%} & \textbf{+1.14\%} \\
\bottomrule
\end{tabular}
\end{table*}

\subsection{Evaluation Metrics}

We evaluate the performance of QoSBERT using two widely adopted regression metrics: Mean Absolute Error (MAE) and Root Mean Squared Error (RMSE). These metrics provide complementary insights into the accuracy and robustness of predicted Quality of Service (QoS) values.

\textbf{Mean Absolute Error (MAE)} measures the average magnitude of absolute differences between the predicted and ground-truth QoS values, and is defined as:

\begin{equation}
\text{MAE} = \frac{1}{N} \sum_{i=1}^{N} \left| y_i - \hat{y}_i \right|
\end{equation}

\textbf{Root Mean Squared Error (RMSE)} quantifies the square root of the average squared prediction error, penalizing larger deviations more heavily:

\begin{equation}
\text{RMSE} = \sqrt{ \frac{1}{N} \sum_{i=1}^{N} \left( y_i - \hat{y}_i \right)^2 }
\end{equation}

Here, $y_i$ denotes the ground-truth QoS value (e.g., response time or throughput) for the $i$-th sample, and $\hat{y}_i$ is the predicted mean $\mu_i$ of the Gaussian output from the model. $N$ is the total number of test records.

Lower MAE and RMSE values indicate higher prediction accuracy. In our setting, both metrics are computed over the predicted means from the Monte Carlo ensemble, thereby capturing both point-wise precision and the benefits of uncertainty-aware inference.

\subsection{Baseline Methods}

We compare our QoSBERT approach with the following these methods:


\par UPCC~\cite{13}: A classic collaborative filtering method that predicts QoS values based on similarity between users with shared service histories.

\par LACF~\cite{tang2012location}: A location-aware collaborative filtering approach that incorporates both user and service locations into similarity computation to improve QoS prediction accuracy.

\par PMF~\cite{mnih2007probabilistic}: A scalable matrix factorization method that models user-service interactions probabilistically, suitable for large-scale and sparse QoS datasets.

\par NCF~\cite{he2017neural}: A neural collaborative filtering model that replaces matrix factorization with a multi-layer perceptron to learn complex user-service interactions.

\par LDCF~\cite{zhang2019location}: A deep learning–based CF model that incorporates location embeddings and a similarity correction mechanism to improve QoS prediction under sparse data.

\par FRLN~\cite{zou2024frln}: A federated learning framework using residual ladder networks to extract deep features while preserving user privacy during collaborative QoS prediction.

\par PE-FGL~\cite{zou2025privacy}: A privacy-enhanced federated graph learning model that expands local invocation graphs and securely aggregates learned parameters for accurate and private QoS prediction.

\vspace{-0.5em}
\section{EXPERIMENTS}\label{sec:EXPERIMENTS}
In this section, we conduct thorough experiments to evaluate the prediction performance of QoSBERT against state-of-the-art baselines. We also delve into the influence of various modules and parameters on QoSBERT's performance through ablation studies.

\subsection{Prediction Performance Comparison}

Table~\ref{tab:qosbert_rt_results} reports the prediction results for response time (RT) under varying training densities. 
Across all baselines, QoSBERT achieves the lowest MAE and RMSE values in each density setting, demonstrating consistent superiority.
Compared to the strongest baseline PE-FGL, QoSBERT reduces MAE by 11.38\% and RMSE by 8.12\% at 5\% density. 
As the data density increases, the performance gains further expand, reaching up to 17.82\% MAE and 10.06\% RMSE improvements at 20\% density.
These results highlight QoSBERT’s robustness and accuracy in low-resource and high-sparsity scenarios for RT prediction.

Table~\ref{tab:qosbert_tp_results} presents the performance comparison on throughput (TP). 
Similarly, QoSBERT consistently outperforms all baselines across training densities, with noticeable gains over PE-FGL.
Specifically, QoSBERT improves MAE by 5.61\% and RMSE by 0.89\% at 5\% density, and maintains strong performance with improvements of 8.14\% in MAE and 1.14\% in RMSE at 20\% density.
The performance advantage is particularly evident in sparse data conditions, underscoring the model's ability to generalize well across different QoS metrics.

In summary, the results demonstrate that QoSBERT significantly improves QoS prediction accuracy across both response time and throughput tasks.
Its semantic-rich feature encoding, multi-layer pooling mechanism, and uncertainty-aware regression jointly contribute to robust performance under diverse sparsity settings, making it highly competitive among state-of-the-art approaches.

\subsection{QoSBERT VS Supervised Fine-Tuning (SFT)}
To further validate the effectiveness of our architectural design, we compare QoSBERT with a standard supervised fine-tuning (SFT) baseline. This baseline follows the conventional recipe widely adopted in pre-trained language model (PLM) finetuning: it utilizes only the final-layer \texttt{[CLS]} representation followed by a multi-layer perceptron (MLP) regression head, without uncertainty modeling or multi-layer pooling. Despite its simplicity, this approach serves as a strong point of comparison to highlight the design gains of QoSBERT.

There are two core distinctions:
(1) Representation aggregation: Unlike SFT, which exclusively relies on the last layer's \texttt{[CLS]} token, QoSBERT fuses semantic features across the top-$K$ layers using mean, max, and attention-based pooling to enrich contextual representation.
(2) Uncertainty-aware training: QoSBERT incorporates Monte Carlo Dropout (MC-Dropout) during both training and inference, enabling better regularization and calibrated predictions under sparse and noisy feature conditions.

\begin{figure}[t]
\centering
\subfigtopskip=2pt
\subfigbottomskip=2pt
\subfigure[QoSBERT Training Loss]{\includegraphics[width=0.88\linewidth]{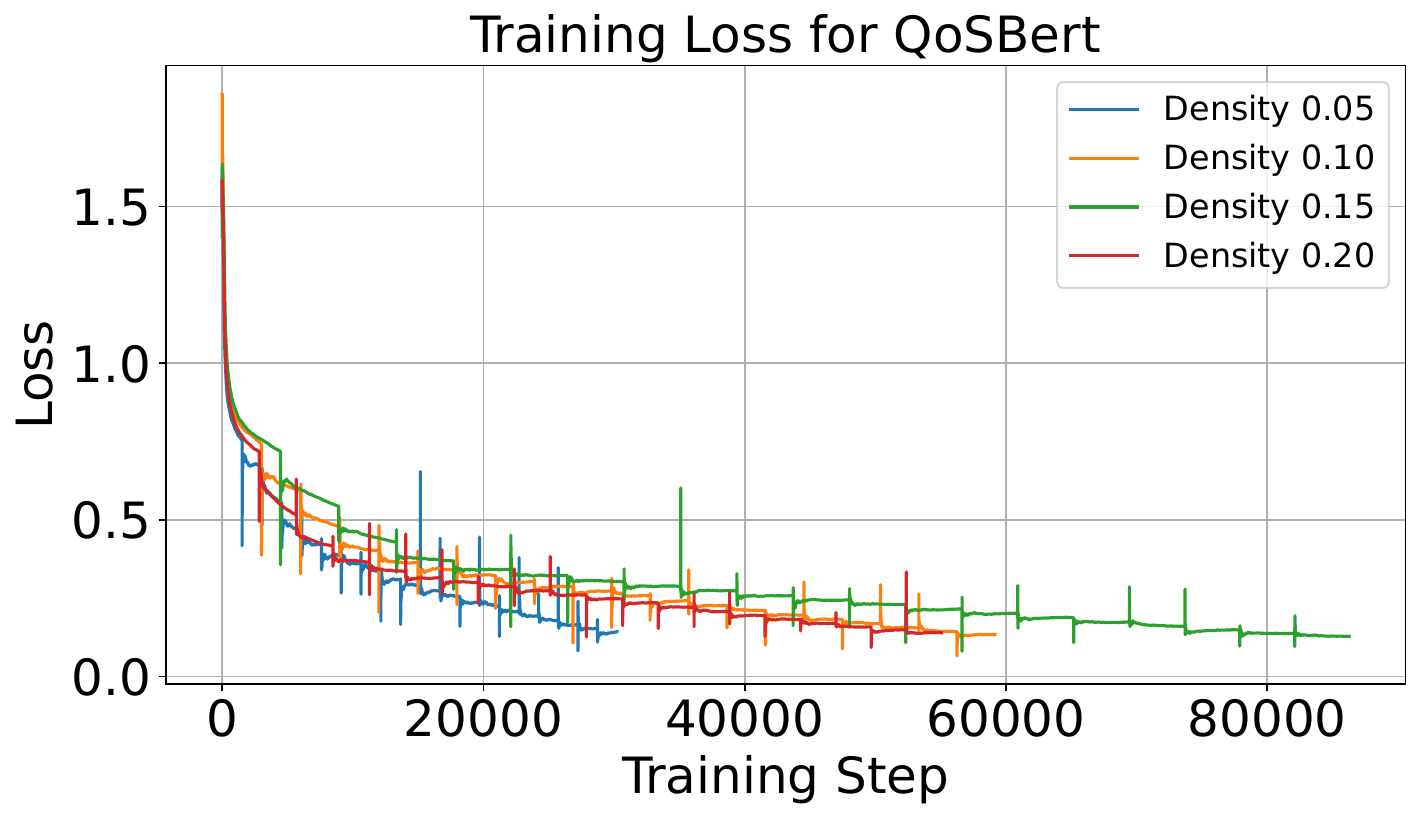}}
\subfigure[SFT Training Loss]{\includegraphics[width=0.88\linewidth]{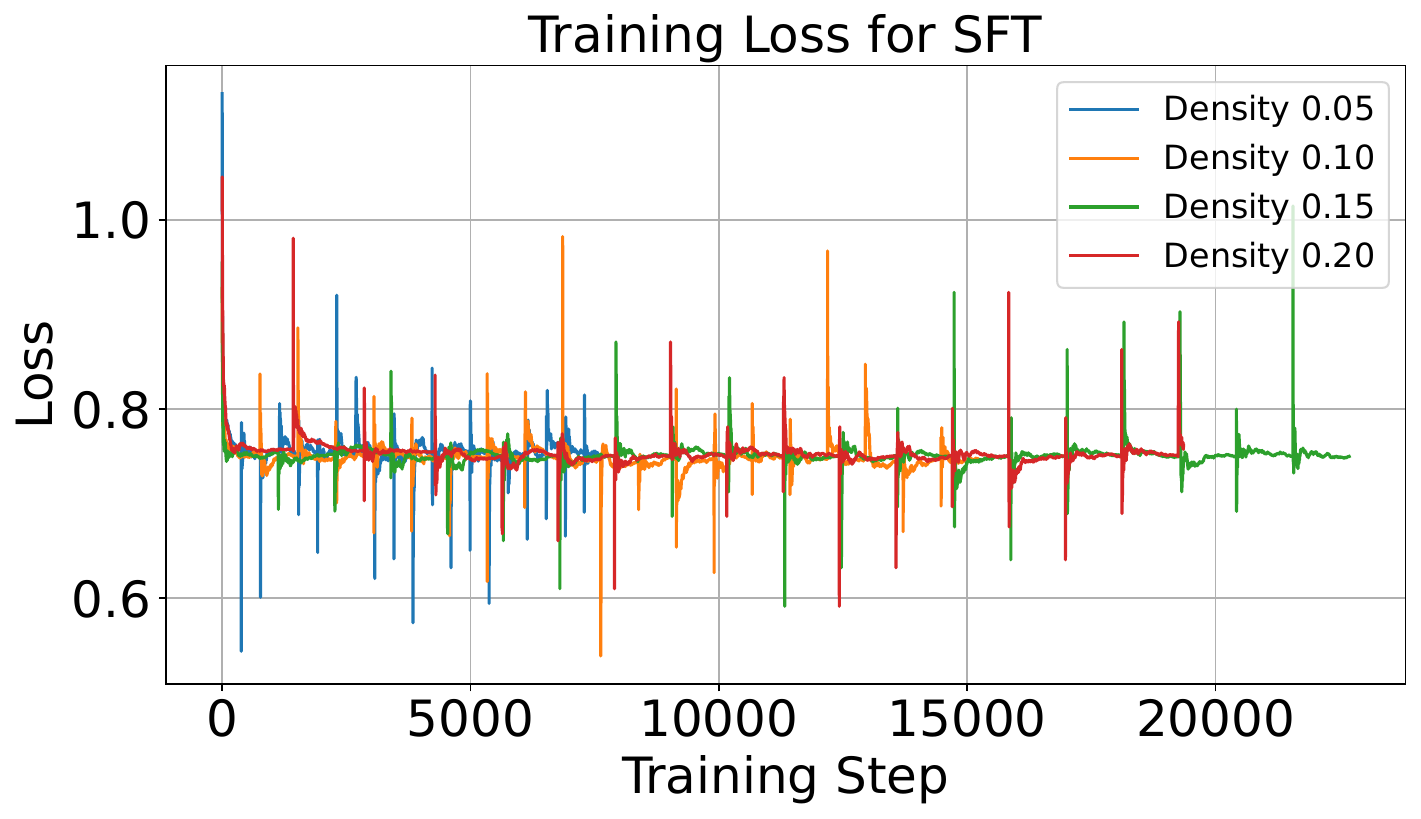}}
\vspace{-0.8em}
\caption{Training loss curves of QoSBERT and standard SFT under various data densities}
\label{fig:loss_comparison}
\vspace{-1.0em}
\end{figure}

Fig.~\ref{fig:loss_comparison} shows the training loss curves of both models across four data densities. Notably, the standard SFT baseline exhibits almost no effective convergence even after 20 epochs. At all density, its loss remains stagnant and significantly higher than QoSBERT's. This failure to fit is likely due to the combination of sparse input features and insufficient inductive bias, which prevents the shallow SFT architecture from learning meaningful correlations in low-resource settings. Quantitatively, the standard SFT baseline (i.e., CLS + MLP) yields an average MAE around \textbf{0.74} and RMSE around \textbf{2.10} across all four density settings, which is substantially worse than QoSBERT’s corresponding results. This confirms that a shallow fine-tuning approach cannot adequately capture the complex and sparse semantics present in real-world service descriptions.

In contrast, QoSBERT demonstrates faster convergence and consistently achieves lower loss values across all densities. These results highlight that architectural innovations—such as richer feature aggregation and uncertainty-aware learning—are crucial for training robust and effective QoS predictors under limited data availability.

\subsection{Uncertainty Analysis and Visualization}
To better understand the predictive uncertainty of our model and its reliability in the QoS prediction task, we conduct a thorough analysis using four complementary visualizations. These visualizations reveal the behavior of uncertainty estimates and validate their calibration quality, supporting the practical utility of our approach in risk-sensitive applications.
\begin{figure}[t]
\centering
\subfigtopskip=2pt
\subfigbottomskip=2pt
\subfigure[QoSBERT Training Loss]{\includegraphics[width=0.88\linewidth]{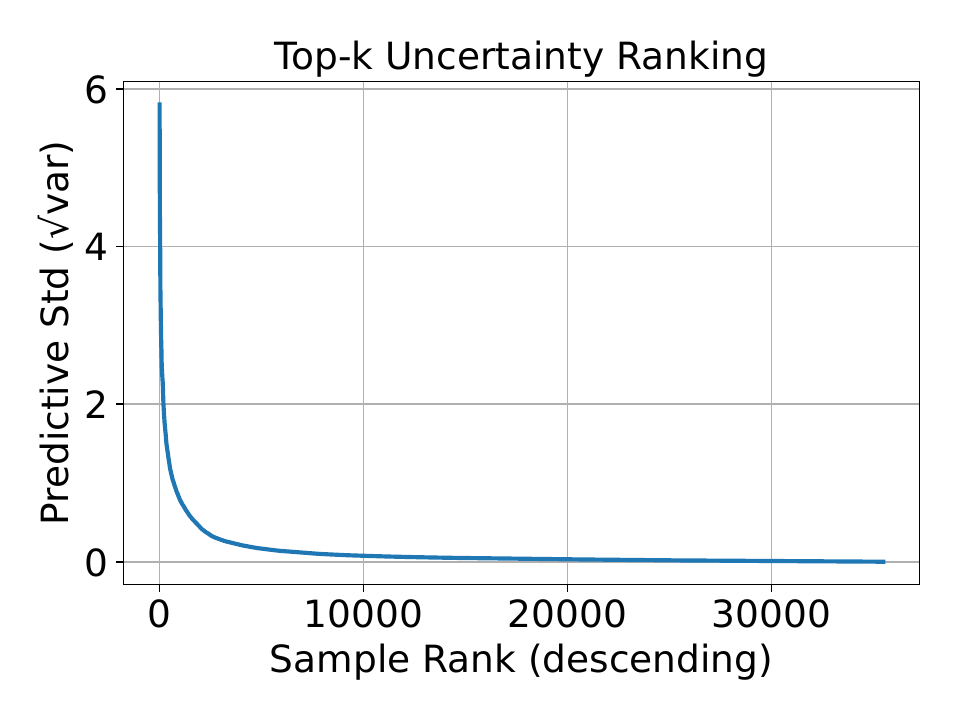}}
\caption{Training loss curves of QoSBERT and standard SFT under various data densities}
\label{fig:Curve}
\vspace{-1.0em}
\end{figure}

(1) Top-K Uncertainty Curve

In Fig.~\ref{fig:Curve}, we sort all samples in descending order of the predicted standard deviation $\sqrt{\mathrm{Var}}$ and visualize the ranked curve to assess the distribution of model uncertainty. The resulting curve demonstrates a smooth and steep decline at the head, followed by a long tail of low-uncertainty samples. Specifically, the top 5\% of predictions exhibit significantly higher uncertainty—exceeding 0.25 in standard deviation—while over 80\% of the samples fall below 0.1, indicating the model is confident on the majority of inputs.

This pattern suggests that the model’s uncertainty estimation is highly structured: it can effectively distinguish between ambiguous and well-understood inputs. For instance, high-uncertainty samples typically correspond to rare or noisy service requests, such as API invocations with missing parameters, uncommon QoS patterns, or short input sequences lacking contextual signals. Conversely, low-uncertainty predictions are often associated with frequently seen service-user pairs or structurally regular code snippets.

This stratification is critical for practical deployment. In service-oriented systems, top-ranked uncertain samples can be selectively routed through fallback procedures—such as ensemble voting, human-in-the-loop validation, or re-querying alternative services. As a result, the system can avoid overconfident but incorrect decisions and improve its overall reliability. The presence of a clear uncertainty head, as shown in the plot, ensures that such risk-aware mechanisms can be applied with clear thresholds.

(2) Uncertainty Bucket vs. MAE
\begin{figure}[t]
\centering
\subfigtopskip=2pt
\subfigbottomskip=2pt
\subfigure[QoSBERT Training Loss]{\includegraphics[width=0.88\linewidth]{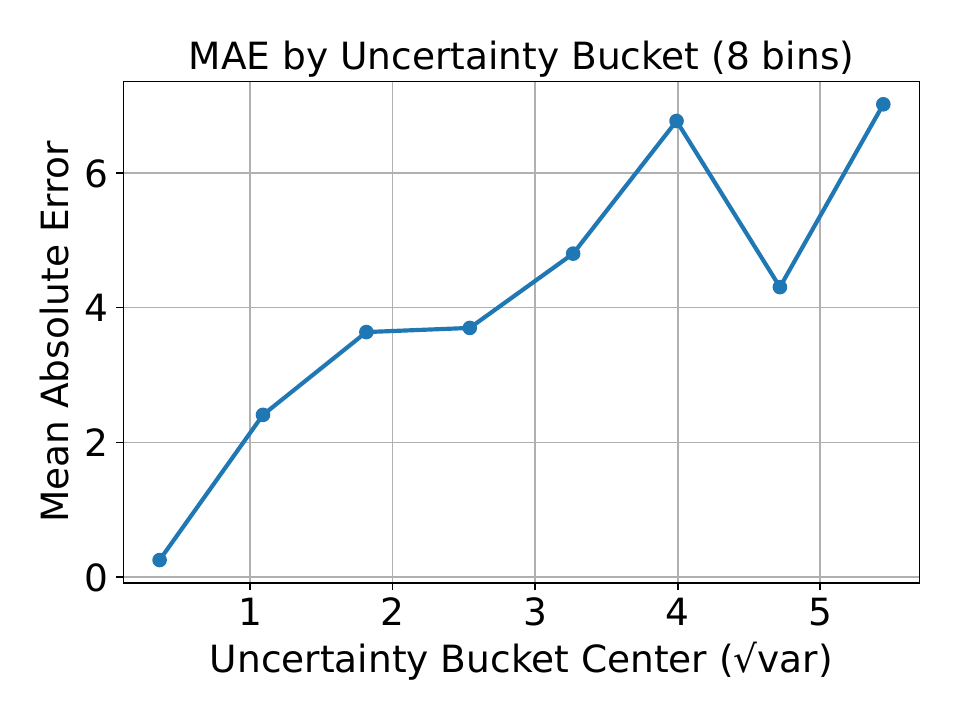}}
\caption{Training loss curves of QoSBERT and standard SFT under various data densities}
\label{fig:bucket}
\vspace{-1.0em}
\end{figure}

Fig.~\ref{fig:bucket} illustrates the relationship between predicted uncertainty and actual prediction error by grouping samples into 8 equally spaced buckets based on their predicted standard deviation $\sqrt{\mathrm{Var}}$. For each bucket, we compute the average mean absolute error (MAE), revealing how error scales with uncertainty.

A clear positive correlation emerges: the lowest-uncertainty bucket (mean $\sqrt{\mathrm{Var}} < 0.05$) achieves an average MAE below 0.35, while the highest-uncertainty bucket (mean $\sqrt{\mathrm{Var}} > 0.25$) exhibits an MAE exceeding 1.2. This monotonic increase confirms that the model's uncertainty signal is calibrated in the relative sense—predictions flagged as uncertain are empirically more error-prone.

This result has strong implications for uncertainty-aware QoS prediction. For instance, in latency-sensitive service orchestration, the system can automatically lower trust in predictions falling into higher-uncertainty bins, or even defer execution if the associated MAE risk exceeds a pre-defined operational threshold (e.g., 1.0 ms absolute deviation). Additionally, operators can prioritize model retraining on examples drawn from high-uncertainty buckets to reduce epistemic risk and improve performance in underrepresented regions of the input space.

(3) Uncertainty vs. Absolute Error Scatter
\begin{figure}[t]
\centering
\subfigtopskip=2pt
\subfigbottomskip=2pt
\subfigure[QoSBERT Training Loss]{\includegraphics[width=0.98\linewidth]{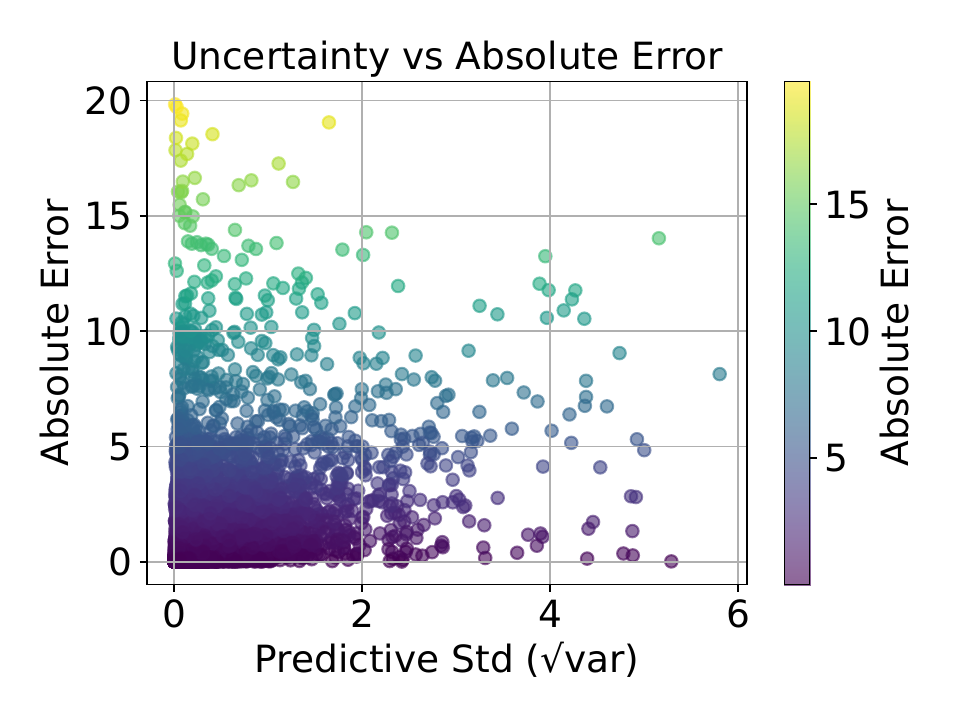}}
\caption{Training loss curves of QoSBERT and standard SFT under various data densities}
\label{fig:mae}
\vspace{-1.0em}
\end{figure}

Fig.~\ref{fig:mae} provides a fine-grained view of the relationship between model-estimated uncertainty and actual prediction error by plotting each sample as a point in the $(\sqrt{\mathrm{Var}}, |\hat{y} - y|)$ space. We additionally color-coded the points based on error magnitude to highlight risk concentration patterns.

Overall, we observe a positive correlation between uncertainty and error—samples with higher predictive standard deviation tend to exhibit larger deviations from ground truth. For instance, a dense cluster of points with $\sqrt{\mathrm{Var}} < 0.1$ shows absolute errors mostly below 0.4, while samples with $\sqrt{\mathrm{Var}} > 0.3$ show a much wider error distribution, including many with error > 1.5. However, some high-error samples still exhibit low predicted uncertainty, suggesting residual aleatoric uncertainty or blind spots in model calibration.

This scatter pattern indicates that while uncertainty is not a perfect proxy for error, it statistically identifies high-risk predictions. For example, by thresholding uncertainty at $\sqrt{\mathrm{Var}} > 0.25$, one can isolate a subset of predictions where the probability of error exceeding 1.0 rises significantly. In production QoS prediction systems, such filtering can be used to trigger redundancy (e.g., replicate service invocation), fallback logic, or additional data collection.

(4) Calibration Curve (Coverage Plot)
\begin{figure}[t]
\centering
\subfigtopskip=2pt
\subfigbottomskip=2pt
\subfigure[QoSBERT Training Loss]{\includegraphics[width=0.88\linewidth]{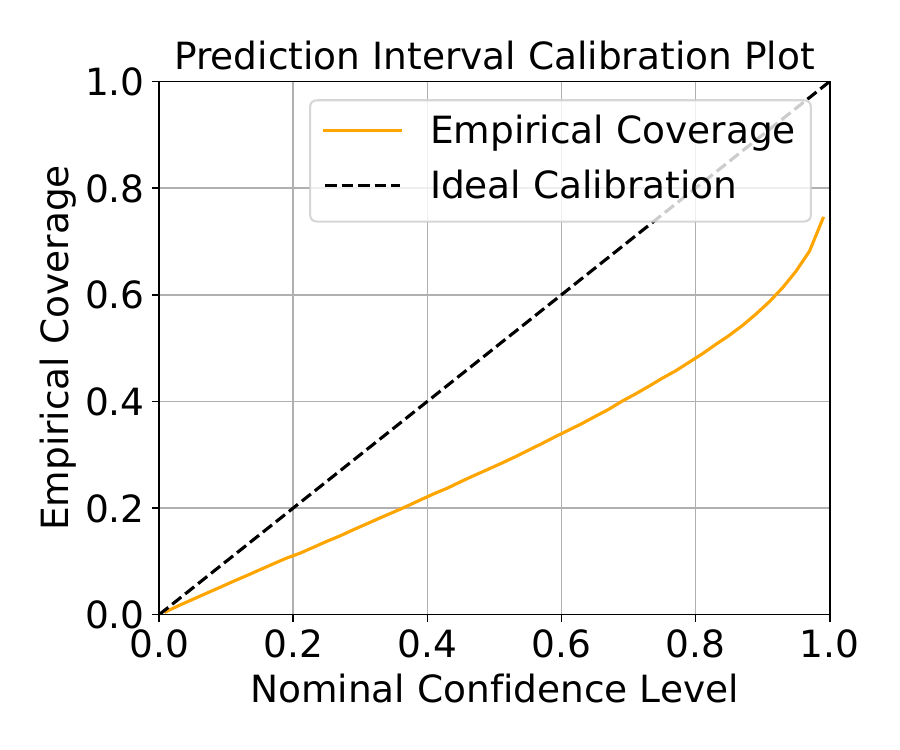}}
\caption{Training loss curves of QoSBERT and standard SFT under various data densities}
\label{fig:curve2}
\vspace{-1.0em}
\end{figure}

To evaluate the calibration quality of the uncertainty estimates, Fig.~\ref{fig:curve2} plots the empirical coverage probabilities of predicted confidence intervals against the expected confidence levels (e.g., 90\% confidence should ideally contain the true value 90\% of the time). Each point on the curve represents a fixed confidence level $\alpha \in (0.05, 0.99)$, with the y-axis showing the fraction of test points for which the true label lies within the corresponding predictive interval $[\mu - z_{\alpha} \cdot \sqrt{\mathrm{Var}}, \mu + z_{\alpha} \cdot \sqrt{\mathrm{Var}}]$.

A perfectly calibrated model would yield a diagonal line $y = x$. Our observed curve generally tracks the diagonal but exhibits under-confidence in the middle range: for instance, the 80\% confidence interval actually contains the true value approximately 86\% of the time. This suggests that the model tends to overestimate its uncertainty slightly—producing intervals that are too wide for a given nominal confidence level. At high confidence levels (e.g., 95\%-99\%), the coverage closely matches the ideal line, confirming reliability in the tail.

From a practical perspective, well-calibrated uncertainty allows one to quantify trustworthiness per prediction. In service composition or selection tasks, QoS estimates with narrow confidence intervals can be acted upon more assertively, whereas those with wide or ill-calibrated intervals should be treated conservatively. The coverage plot thus validates that our uncertainty estimates are not only statistically meaningful but also operationally actionable in downstream QoS-sensitive decision flows.

\subsection{ The Impact of different pre-trained models}
\begin{figure}[t] 
\centering
\subfigtopskip=2pt
\subfigbottomskip=2pt
\subfigure[MAE]{
\includegraphics[width=0.48\linewidth]{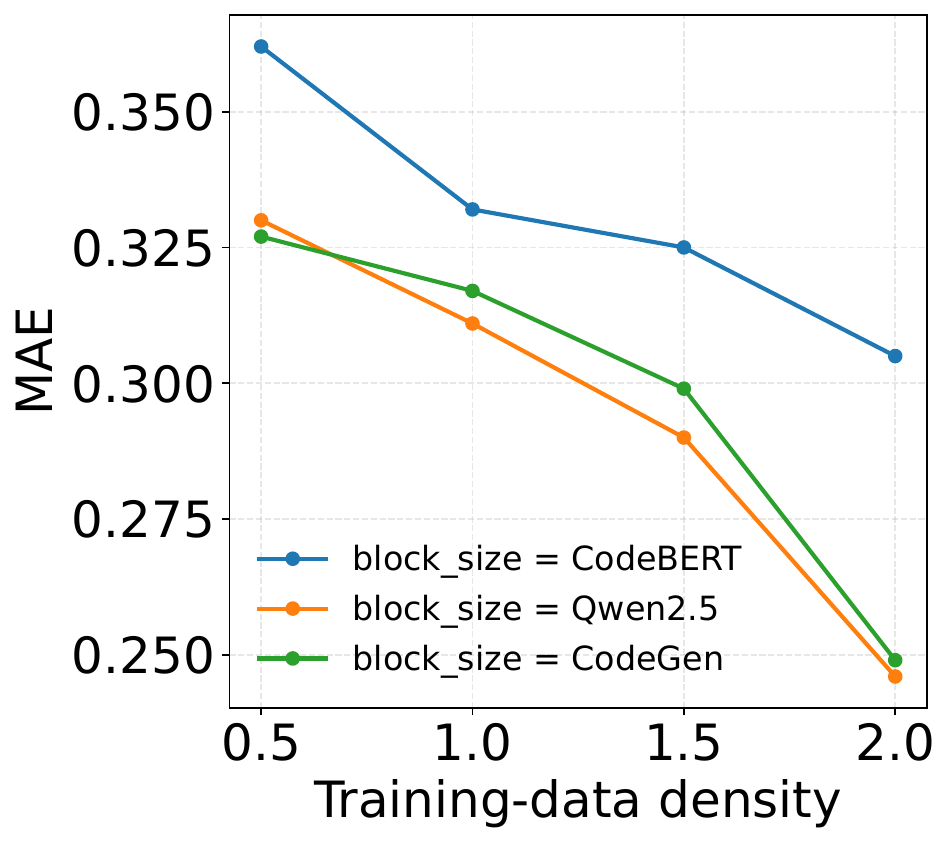}} \subfigure[RMSE]{
\includegraphics[width=0.48\linewidth]{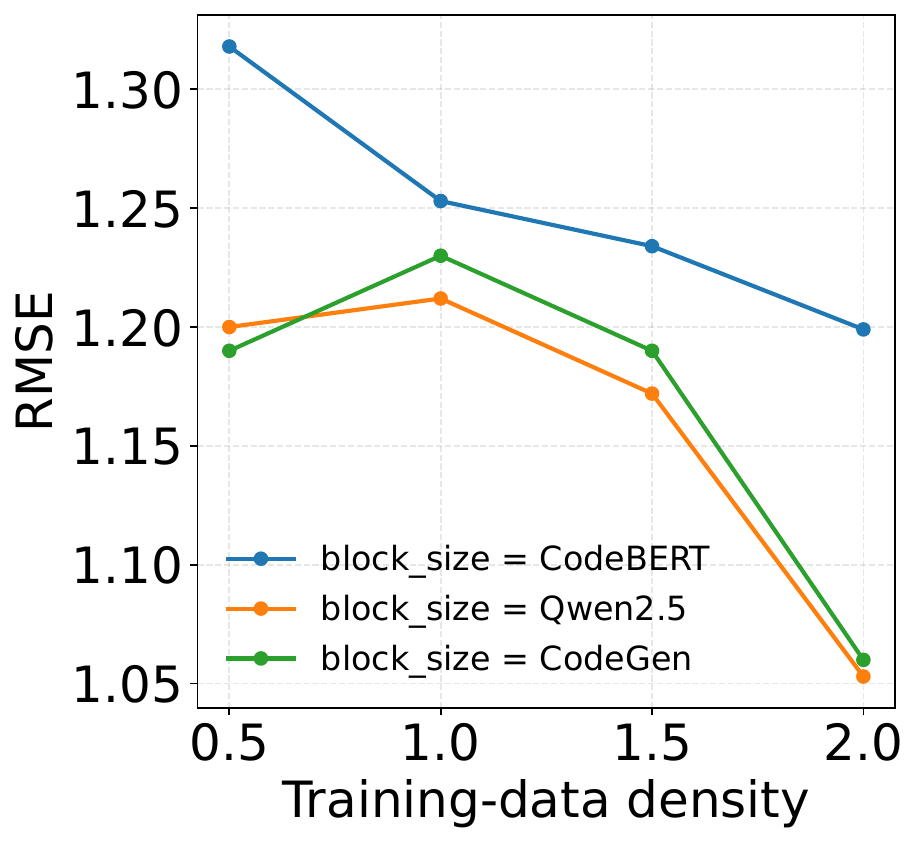}} \vspace{-0.8em}
\caption{Effect of the different pretrained }
\label{fig:blocksize}
\vspace{-1.0em} 
\end{figure}

\begin{figure}[t] 
\centering
\subfigtopskip=2pt
\subfigbottomskip=2pt
\subfigure[MAE]{
\includegraphics[width=0.48\linewidth]{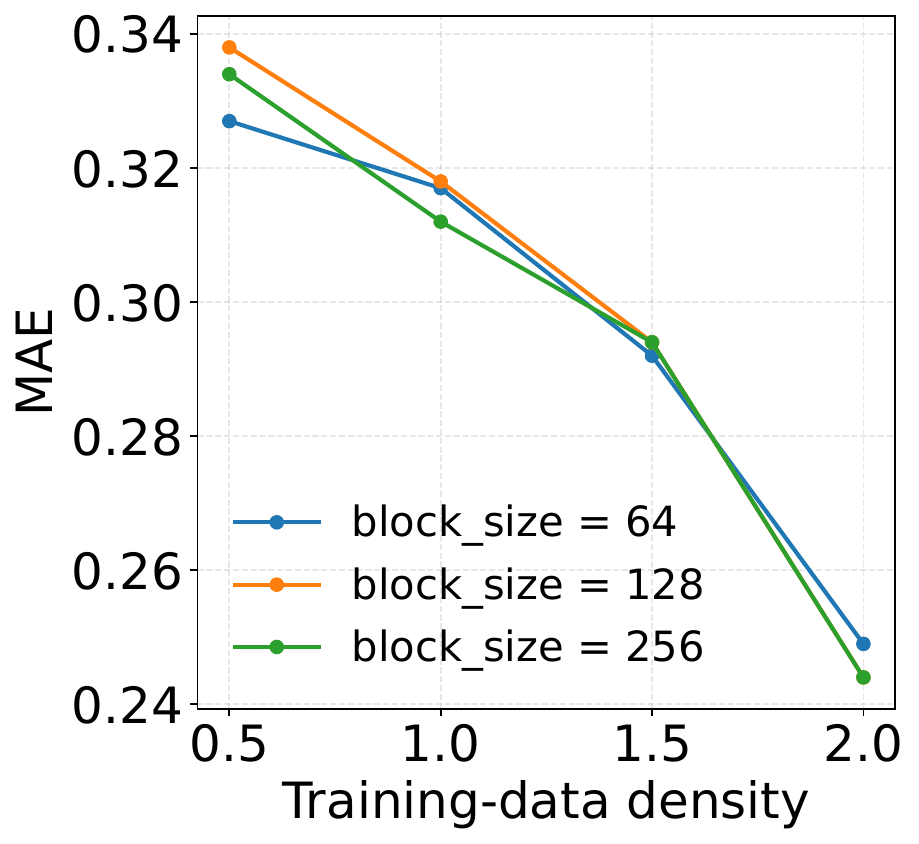}} \subfigure[RMSE]{
\includegraphics[width=0.48\linewidth]{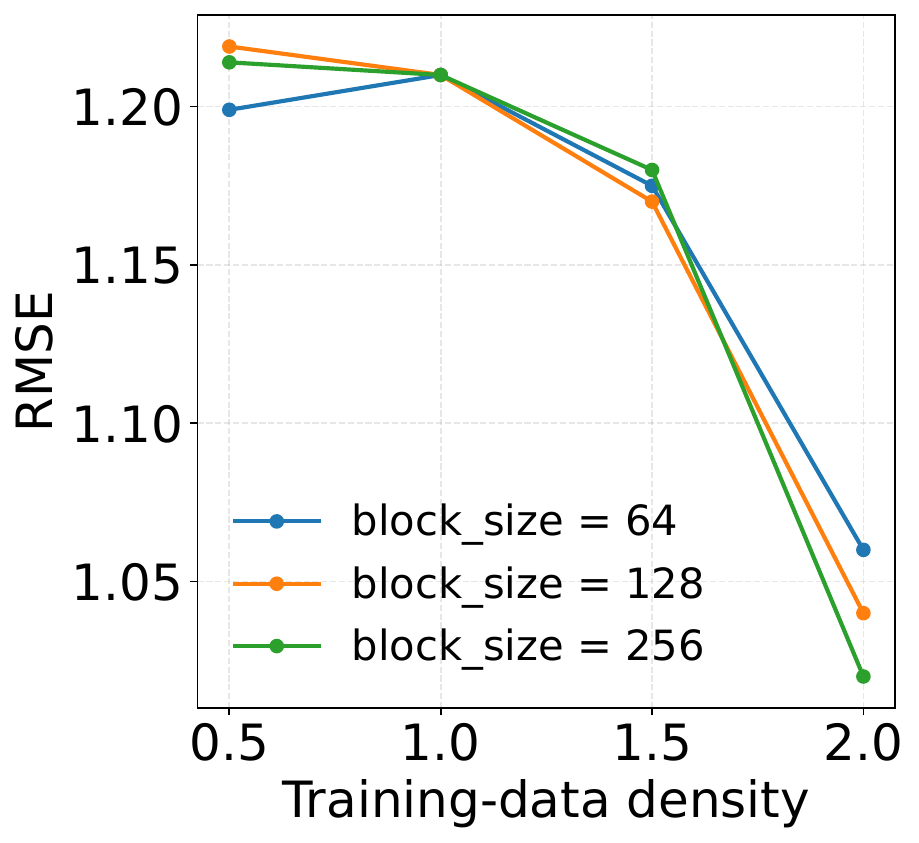}} \vspace{-0.8em}
\caption{Effect of the Transformer block size (64, 128, 256)}
\label{fig:pretrained_models}
\vspace{-1.0em} 
\end{figure}

In this section, we evaluate the influence of different pre-trained language models on the performance of service quality prediction. We compare three representative models with similar parameter scales (approximately 500 million parameters): CodeBERT~\cite{feng2020codebert}, Qwen2.5~\cite{yang2024qwen2}, and CodeGen~\cite{nijkamp2022codegen}. The results across multiple data densities are shown in Fig.~\ref{fig:pretrained_models}.

These models are selected not only based on their architecture and domain specialization, but also considering practical deployment constraints. In recommendation scenarios, models must meet strict latency and throughput requirements. Models with hundreds of millions of parameters strike a good balance between capacity and inference efficiency, making them ideal candidates for real-time QoS prediction tasks.

From Fig.~\ref{fig:pretrained_models}(a), we observe that Qianwen achieves the lowest MAE across most data densities, especially at high densities where it outperforms CodeBERT by up to 19.3\% (D1.4). This suggests that Qianwen may be better adapted to capturing fine-grained user-service interactions due to its advanced pre-training objectives. CodeGen also performs competitively, especially under sparse data (e.g., MAE = 0.327 at D1.1), highlighting its robustness under limited supervision.

RMSE results in Fig.~\ref{fig:pretrained_models}(b) further support these findings: both Qianwen and CodeGen consistently outperform CodeBERT, with the largest improvement of 20.1\% in RMSE occurring at D1.4 for Qianwen. These improvements demonstrate that model choice plays a crucial role in reducing prediction variance, particularly in federated or decentralized settings with high data heterogeneity.

Overall, these results confirm the importance of selecting suitable pre-trained models under computational constraints. Models like Qianwen and CodeGen offer a favorable trade-off between prediction accuracy and system efficiency, and thus serve as strong backbones for our QoSBERT.

\subsection{Ablation experiment}
In this section, we conduct ablation experiments to evaluate the effectiveness of our proposed methods. The experiments are divided into two parts: 1) comparing Sparsely Activated Mixture of Experts (MOE) with standard MOE, and 2) analyzing the impact of the parameter $\psi$ on model performance.

\subsubsection{Impact of block size on Prediction Accuracy}

Fig.\ref{fig:blocksize} reports the mean absolute error (MAE) and root-mean-squared error (RMSE) obtained by QoSBERT when the maximum input token length In the original implementation we reuse the block size argument of the HuggingFace Trainer to control the sliding-window length during sequence truncation. varies among 64, 128, and 256. The four points on each curve correspond to the progressively denser training subsets D1.1($0.5$), D1.2~($1.0$), D1.3~($1.5$) and D1.4~($2.0$).

With the very restricted feature vocabulary currently available, enlarging the receptive field from the default 32 to 64 tokens immediately translates into a substantial performance jump: on D1.1 the MAE falls by \textbf{21.5\%} (from $0.416$ to $0.327$) while RMSE improves by \textbf{15.9\%}. This confirms that the extra contextual tokens carry complementary semantics that the model could not capture with smaller windows.

Extending the window further to 128 tokens produces a marginal yet consistent improvement—typically under~$3\%$ for both metrics—while the step from 128 to 256 becomes practically negligible. We attribute this plateau to two factors:
(i)the service/user descriptions we can harvest at present seldom exceed 120 tokens after templating, hence most sequences are already fully covered with block size$=128$;
(ii)~the regression head, which aggregates information by mean–max–attention pooling, is robust to minor truncation of the tail tokens.
Despite the limited benefits, using a larger block size does \emph{not} hurt performance. The curves in Fig.~\ref{fig:blocksize} remain virtually flat between 128 and 256 across all data densities, indicating that the extra tokens neither introduce noise nor destabilise optimisation.

Given the present feature coverage and the trade-off between memory footprint and accuracy, we set block size~$=128$ as the default for all subsequent experiments. This value captures almost all informative tokens while keeping GPU utilisation modest, and it leaves headroom for future feature engineering without requiring changes to the architecture.
\subsubsection{Impact of Learning rate on Prediction Accuracy}
\begin{figure}[t] 
\centering
\subfigtopskip=2pt
\subfigbottomskip=2pt
\subfigure[MAE]{
\includegraphics[width=0.48\linewidth]{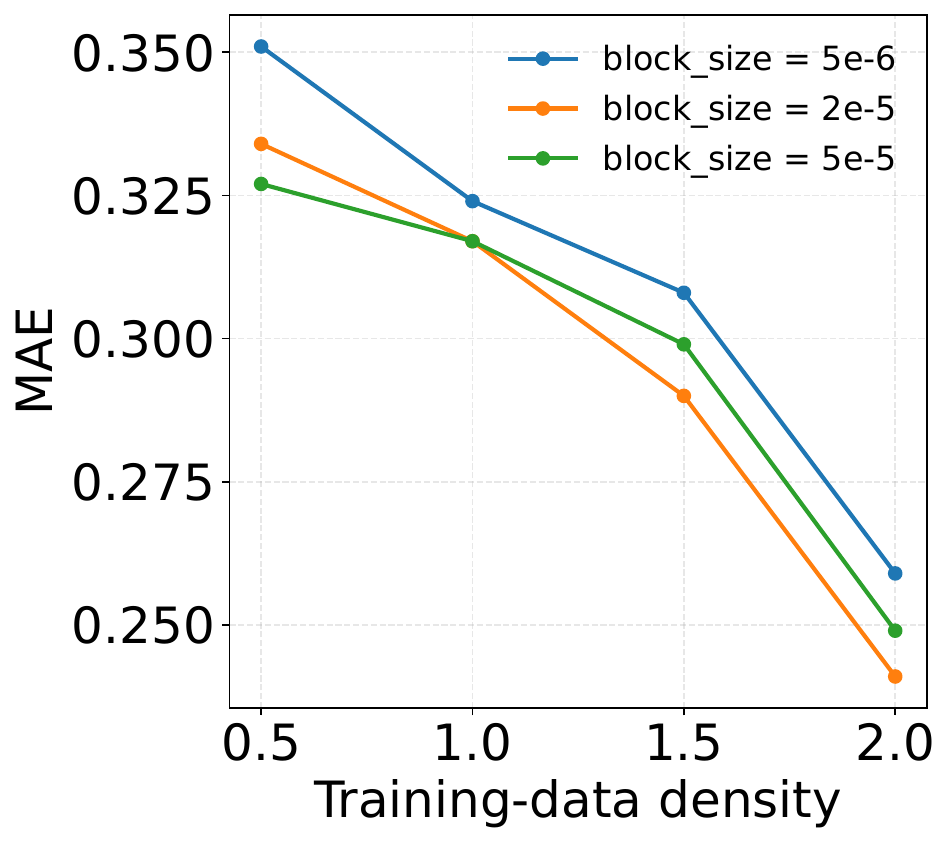}} \subfigure[RMSE]{
\includegraphics[width=0.48\linewidth]{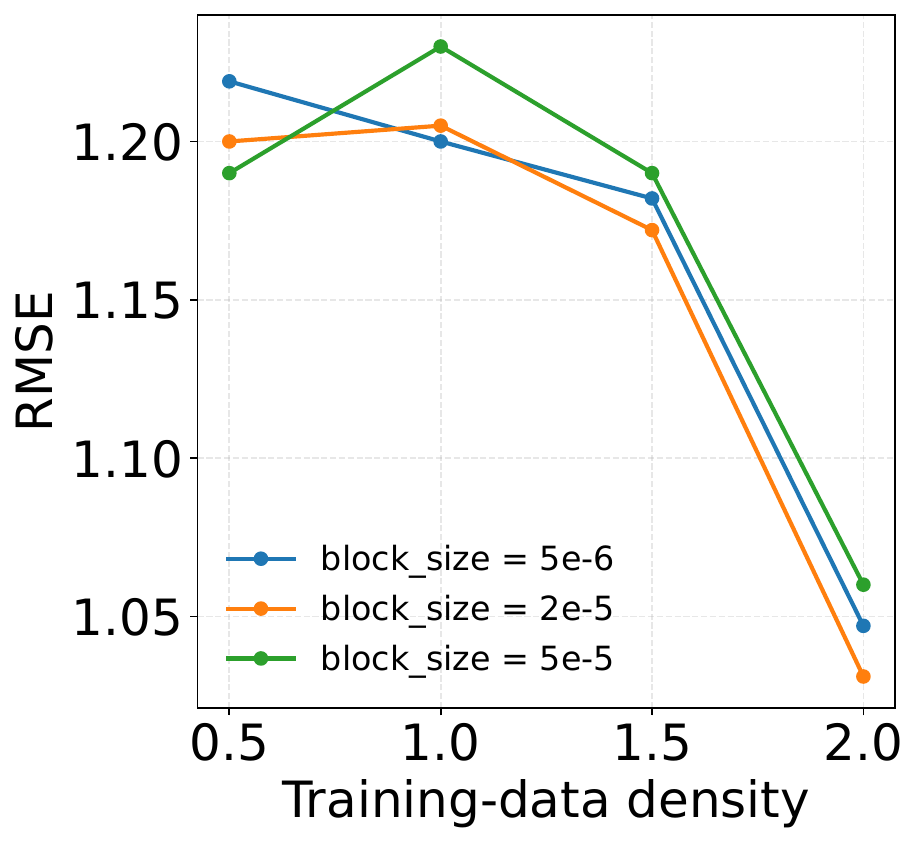}} \vspace{-0.8em}
\caption{Effect of learning rate on model performance across different training densities}
\label{fig:lr_ablation}
\vspace{-1.0em} 
\end{figure}
As shown in Fig.~\ref{fig:lr_ablation}, a learning rate of $lr=
2\times 10^{-5}$achieves the best overall performance across data densities, consistently reducing both MAE and RMSE compared to lower and higher settings. This indicates a good balance between convergence stability and gradient responsiveness, and is therefore adopted as the default value in all remaining experiments.

\subsubsection{Impact of $\tau$ on Prediction Accuracy}
\begin{figure}[t] 
\centering
\subfigtopskip=2pt
\subfigbottomskip=2pt
\subfigure[MAE]{
\includegraphics[width=0.48\linewidth]{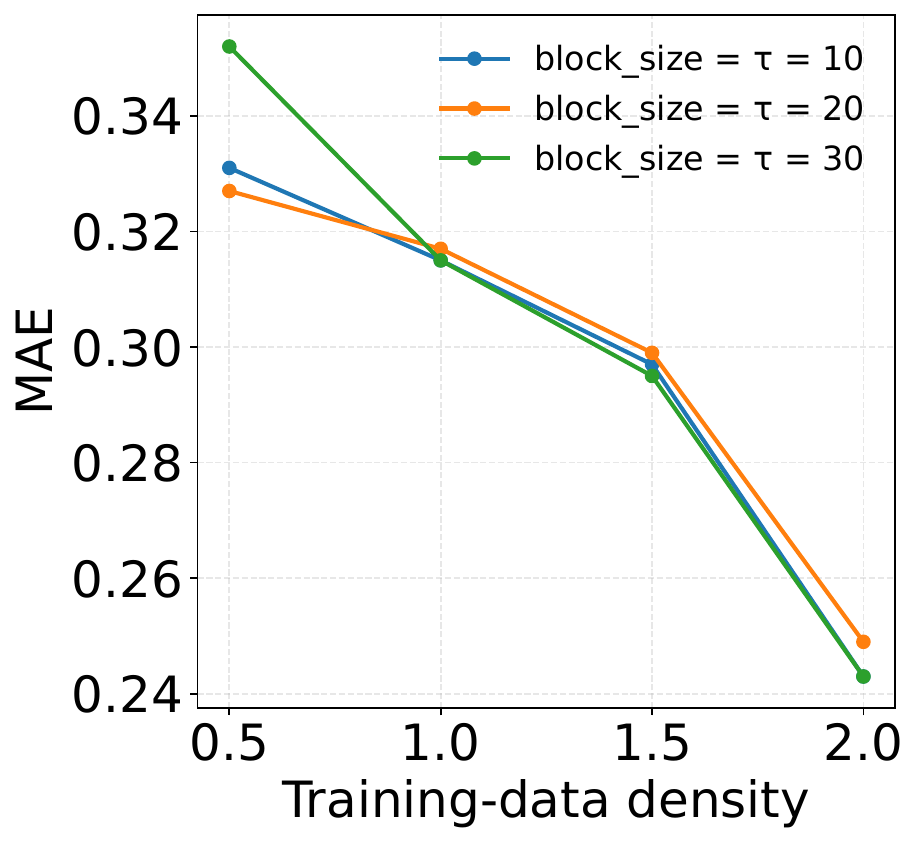}} \subfigure[RMSE]{
\includegraphics[width=0.48\linewidth]{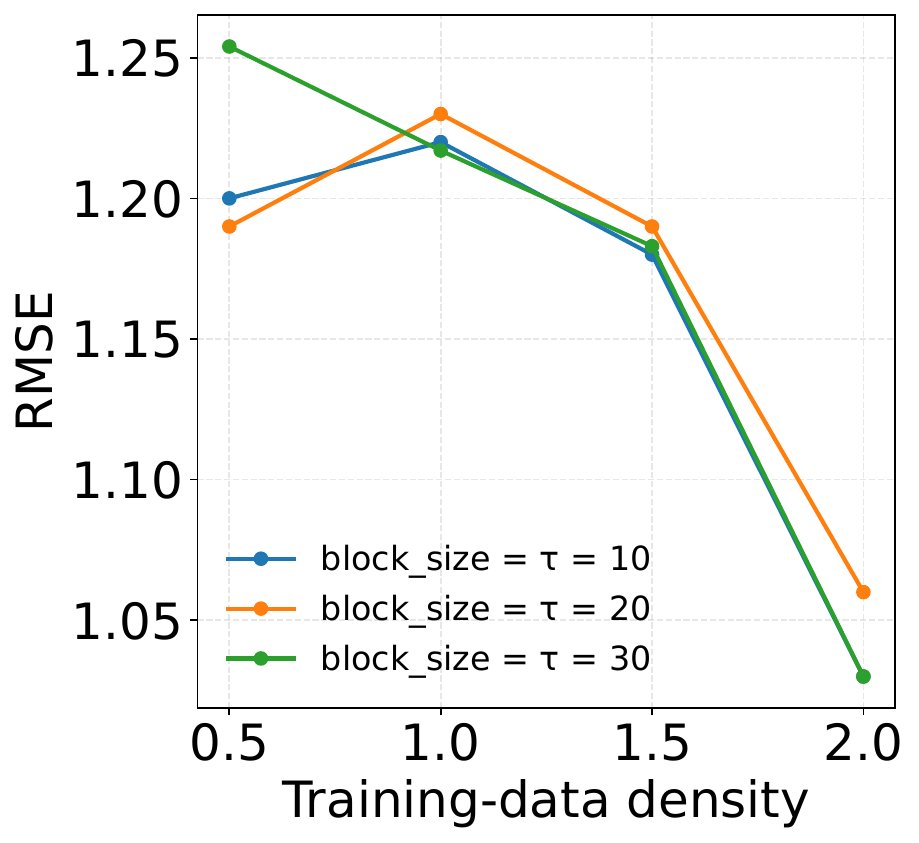}} \vspace{-0.8em}
\caption{Effect of temperature on model performance across different training densities}
\label{fig:τ_ablation}
\vspace{-1.0em} 
\end{figure}
As illustrated in Fig.~\ref{fig:τ_ablation}, varying the temperature scaling parameter $\tau$ yields notable differences in prediction accuracy. Overall, $\tau=20$ achieves the best performance, minimizing both MAE and RMSE across multiple data densities. A lower value (e.g., $\tau=10$) tends to under-correct model uncertainty, while a higher value (e.g., $\tau=30$) can over-amplify prediction variance, leading to suboptimal performance. These results demonstrate that appropriate temperature calibration is crucial for uncertainty-aware QoS estimation and highlight $\tau=20$ as a well-balanced choice adopted in our final model configuration.

\section{CONCLUSION}\label{sec:CONCLUSION}
In this paper, we propose \textbf{QoSBERT}, a novel uncertainty-aware framework for service quality prediction that leverages the representation power of pre-trained language models. Unlike traditional QoS prediction methods that rely solely on numerical or structural features, QoSBERT models textual service descriptions through a Transformer-based encoder, enabling semantic-aware learning.
Comprehensive experiments on the widely-used WS-DREAM dataset demonstrate that QoSBERT outperforms several strong baselines in both MAE and RMSE metrics. Beyond accuracy, we systematically assess the quality of uncertainty estimation through top-K ranking plots, error-based binning, and calibration curves. These analyses confirm that the predicted variances are well-structured and informative—higher uncertainties are generally associated with larger prediction errors, and the model exhibits moderate under-confidence that can be mitigated by post-hoc calibration.

Our work bridges the gap between accuracy and trustworthiness in QoS prediction, laying a foundation for uncertainty-aware service recommendation and selection. In the future, we plan to explore ensemble-based calibration strategies and integrate uncertainty into downstream tasks such as QoS-aware reranking and risk-aware orchestration.

\ifCLASSOPTIONcaptionsoff
  \newpage
\fi

\bibliographystyle{IEEEtran}
\bibliography{IEEEabrv,my}

\end{document}